\documentclass{article}
\usepackage[preprint,nonatbib]{neurips_2020}

\usepackage[utf8]{inputenc} 
\usepackage[T1]{fontenc}    
\usepackage{hyperref}       
\usepackage{url}            
\usepackage{booktabs}       
\usepackage{amsfonts}       
\usepackage{nicefrac}       
\usepackage{microtype}      

\usepackage{amsmath,bm}
\usepackage{graphicx}
\usepackage{dcolumn}
\usepackage{subcaption,soul}
\usepackage{caption}
\usepackage[inline]{enumitem}
\usepackage[numbers,sort&compress,sectionbib]{natbib}
\usepackage{mleftright}
\usepackage{multirow}

\usepackage{titlesec}
\usepackage[redeflists]{IEEEtrantools}

\title{COVID-Net Assistant: A Deep Learning-Driven Virtual Assistant for COVID-19 Symptom Prediction and Recommendation}

\author{%
  Pengyuan Shi$^{1\dagger}$, Yuetong Wang$^{1\dagger}$, Saad Abbasi$^{2}$, Alexander Wong$^{2}$\\
  $^{1}$David R.\ Cheriton School of Computer Science\\
  $^{2}$Department of Systems Design Engineering\\  
  University of Waterloo\\
}

\begin{document}
\maketitle
\def\thefootnote{$\dagger$}\footnotetext{Denotes equal contribution.}\def\thefootnote{\arabic{footnote}}

\begin{abstract}
As the COVID-19 pandemic continues to put a significant burden on healthcare systems worldwide, there has been growing interest in finding inexpensive symptom pre-screening and recommendation methods to assist in efficiently using available medical resources such as PCR tests. In this study, we introduce the design of COVID-Net Assistant, an efficient virtual assistant designed to provide symptom prediction and recommendations for COVID-19 by analyzing users' cough recordings through deep convolutional neural networks. We explore a variety of highly customized, lightweight convolutional neural network architectures generated via machine-driven design exploration (which we refer to as COVID-Net Assistant neural networks) on the Covid19-Cough benchmark dataset. The Covid19-Cough dataset comprises 682 cough recordings from a COVID-19 positive cohort and 642 from a COVID-19 negative cohort. Among the 682 cough recordings labeled positive, 382 recordings were verified by PCR test. Our experimental results show promising, with the COVID-Net Assistant neural networks demonstrating robust predictive performance, achieving AUC scores of over 0.93, with the best score over 0.95 while being fast and efficient in inference. The COVID-Net Assistant models are made available in an open source manner through the COVID-Net open initiative and, while not a production-ready solution, we hope their availability acts as a good resource for clinical scientists, machine learning researchers, as well as citizen scientists to develop innovative solutions. 

\end{abstract}

\section{Introduction}

The coronavirus 2019 (COVID-19) pandemic, caused by severe acute respiratory syndrome coronavirus 2 (SARS-CoV-2), continues to impact human society worldwide significantly. Real-time reverse transcription polymerase chain reaction (RT-PCR) testing remains the most reliable standard screening tool for detecting COVID-19. While RT-PCR testing is the primary screening tool against COVID-19, it typically requires two to three days to complete and significantly burdens the healthcare system. Another widely used screening test is the antigen test. They are rapid tests that produce results in 15-30 minutes. While medical screening tests are becoming more accessible and widespread, many countries do not have widespread access, especially in some developing countries. 

Earlier works show deep learning is successful on particular COVID-19  classification tasks~\cite{covidnet, covidnet-ct} with deep convolutional neural networks (CNNs). Deep learning algorithms involve using a dataset to train a tailored parameterized model to help detect symptoms and scientific evidence. The success of deep learning led researchers to explore new diagnosis methods via visual information, such as chest X-rays and CT images. However, a successful deep learning algorithm usually leverages a large dataset with high-quality labels, which is expensive in medical studies. Nevertheless, such visual information is challenging to retrieve in a short time. A more affordable and efficient pre-screening method is beneficial to the use of medical resources.

Coughing is one of the primary symptoms of COVID-19 and is also the consequence of other diseases. Due to the difference in infection sources, cough sounds may differ in patterns hardly detectable by humans. With signal processing techniques, imperceptible patterns can be learned and detected by deep learning techniques. One of the signal processing techniques is Mel-frequency cepstral coefficients (MFCCs), which work well with CNNs.

\begin{figure}[t]
    \vspace{20pt}	    
	\begin{center}
		\includegraphics[scale = 0.3]{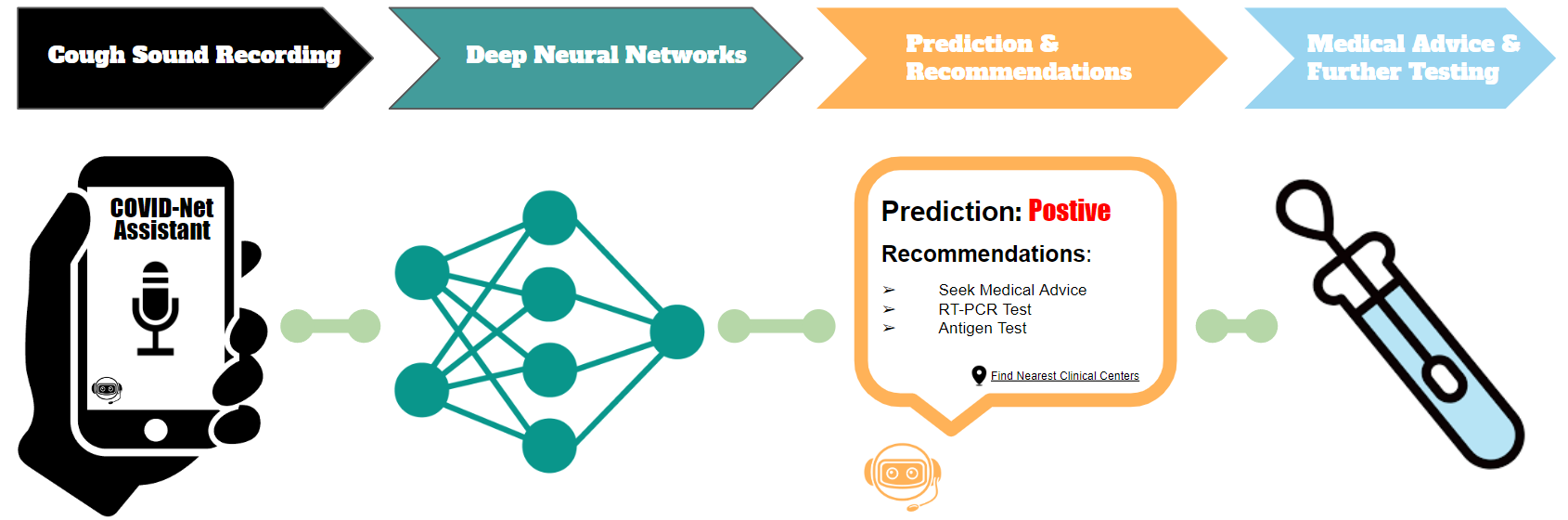}
		\caption{ Overview of COVID-Net Assistant Workflow}
		\label{app-workflow}
	\end{center}
	\vspace{-10pt}	
	
\end{figure}

Motivated by providing an affordable and accessible tool to provide early recommendations and leverage the use of diagnostic tests in the fight against the COVID-19 pandemic, in this work, we present a design of a virtual assistant \textrm{---} COVID-Net Assistant. It can sit on a mobile device or website to provide early COVID-19 recommendations based on cough recordings due to the feasibility of retrieval. Figure~\ref{app-workflow} shows the workflow of COVID-Net Assistant:
\begin{enumerate*}[label={\arabic*)}]
	\item An individual records or uploads cough audio through the application;
	\item The audio gets processed by an efficient deep convolutions neural network to predict whether the individual may have signs of COVID-19;
	\item If the network predicts signs of COVID-19 based on the audio, it will give the user a recommendation that they may have signs and seek medical advice or further testing using COVID-19 tests such as PCR and antigen tests.
\end{enumerate*}

We hope the open-source nature of the COVID-Net Assistant \footnote{https://github.com/fshipy/COVIDNet-Assistant} encourages further innovation as part of the COVID-Net global open initiative \footnote{http://www.covid-net.ml/}

\section{Data}

We found three public cough datasets that frequently used in relevant research: the COUGHVID~\cite{coughvid} dataset, Coswara~\cite{coswara} dataset, and Covid19-Cough~\cite{covid19-cough} dataset. The positive samples in COUGHVID and Coswara datasets are all self-reported, which does not guarantee the correctness of labels. In our study, we require strong labels to indicate the status of COVID-19 infection; therefore, we chose Covid19-Cough as the main dataset to move forward.

This Covid19-Cough dataset consists of 1324 raw audio samples with 682 cases labeled positive, and 382 of them are confirmed by PCR test. For the audio pieces labeled negative, no verification information is provided. The audio was collected through a call center or telegram. We excluded two malformed audio files in our study, leaving us with 1322 pieces of audio, with 681 labeled positive and 381 positive samples verified.

In our study, we trained deep learning models with two dataset splits. The first split, namely Verified-Only, excludes all unverified positive samples as we aimed to experiment with high-quality labels. The other split, namely All-Data, contains a total of 1322 valid samples from the  Covid19-Cough dataset. We used 60\%, 20\%, and 20\% of the audio for each dataset split as the training, validation, and testing data, respectively. Table~\ref{data-table} shows the exact label distribution in each dataset split.

\begin{table}[ht]
	\centering
    \caption{Dataset Split and Label Distribution.}
	\label{data-table}
    \vspace{0.1cm}
    \begin{tabular}{l|l|l|l}
    \toprule
	Dataset Split&Sub-split&	Positive & Negative	\\ 
    \midrule
	All-Data & Train & 424  &  369\\
	& Validation & 138   &  126\\
	& Test & 119   &  146\\
    \midrule
     Verified-Only & Train & 238  &  375\\
     & Validation & 73   &  131\\
     & Test & 70   &  135\\
    \bottomrule
    \end{tabular}
\end{table}

\section{Methodology}

\subsection{Data Processing}

To leverage the power of deep convolutional neural networks (CNNs), we processed the raw audio using the librosa~\cite{librosa} library to extract  Mel-frequency cepstral coefficients (MFCC) features. MFCC is widely used in audio processing and is used in related work \cite{relate-work2, relate-work1}.
The complete data processing pipeline comprises the following steps:

\begin{enumerate}[]
	\item Load raw audio using librosa~\cite{librosa}.
	\item Apply data augmentation to raw audio.
	\item Extract MFCC features from augmented audio.
\end{enumerate}
The raw audio files are loaded with a sample rate of 48000 Hz. We applied data augmentation to alleviate the scarcity of data and the over-fitting issue, as we will address later. We found that data augmentation helps to scale the data and improve model performance. We used audiomentations \footnote{https://github.com/iver56/audiomentations}, another python library, to apply the augmentations. For every piece of training audio, another five augmented pieces are generated with random trimming, shifting, Gaussian noises, and pitch shifting. 

Afterward, MFCC features are extracted using librosa~\cite{librosa}. By using MFCC, we hope it can reveal strong signals in the data and filter out noisy signals.
To extract the MFCC features from audio, the time domain input is first mapped to the frequency domain by taking the discrete Fourier transform.
Then a mel-scaling, defined by Eq.~\ref{eq:1}, is applied to the frequency domain data to map the frequencies to conceptually equally distanced pitches as perceived by humans.

\begin{equation}
m = 2595 * \log(1 + \frac{f}{100})\label{eq:1}
\end{equation}
Finally, a discrete cosine transform is applied to the log of the previous result to obtain the coefficients. In our work, we used 32 MFCC coefficients, 32 mel bands, and an FFT window size of 2048. The MFCC feature extraction will transform an audio into an image-like data with shape $ 32\times 328\times 1$ to be used as the input to deep convolutional neural networks. Compared to related works \cite{relate-work2}, we used more MFCC coefficients for two reasons. First, we believe strong signals in cough sounds might be hidden in information imperceptible by humans, encoded by higher MFCC coefficients. Second, we believe that by using generative synthesis~\cite{gensynth}, we will be able to train and optimize the models that are aware of which part of the data is essential.

\subsection{Machine-driven Design Exploration}
Inspired from earlier works of COVID-Nets~\cite{covidnet, covidnet-ct}, the final architecture designs of deep neural networks in COVID-Net Assistant were discovered automatically via a machine-driven design exploration process using generative synthesis~\cite{gensynth}. The architecture exploration process identified the optimal macroarchitecture and microarchitecture designs of the tailored model architecture.  With user-defined performance metrics, dataset, and seed architecture, the optimal architecture is determined via an iterative constrained optimization process based on a universal performance function (e.g.,~\cite{netscore}) and a set of quantitative constraints. The architectures discovered via generative synthesis are highly customized designs that compromise complexity and representational power, that is outperforming manual design with greater flexibility and granularity~\cite{covidnet}.

\subsection{Model Architectures} \label{model}  
To leverage the power of generative synthesis~\cite{gensynth}, we built the seed architectures by experimenting with three fundamental convolution-based cells, which are prevalent in building light-weight computer vision models: 
\begin{enumerate*}[label={\arabic*)}]
	\item COVID-Net Assistant CNN built with standard convolutions;
	\item COVID-Net Assistant Res-CNN build with residual blocks~\cite{resnet};
	\item  COVID-Net Assistant DW-CNN build with depth-wise separable convolutions~\cite{mobilenetv1}.
\end{enumerate*}
In particular, residual blocks bring benefits in building deep networks, as it uses identity mapping to tackle the vanishing gradient problem~\cite{res-beni}. Depth-wise separable convolutions increase the representation power of a model while reducing the number of parameters and computation~\cite{dw-conv}.  

Due to the limited training data, we found that the originally proposed ResNet~\cite{resnet} and MobileNetV1~\cite{mobilenetv1} architectures can easily over-fit the training set, and result in a poor performance in the test set. To tackle the over-fitting issue, we built models with dropout layers and fewer parameters. In our experiments, we built multiple seed designs with the aforementioned fundamental cells, as shown in Table~\ref{archetecture}. For each model architecture built with residual blocks (Res) and depth-wise separable convolutions (DW), we built three variants (S, M, L) with increasing complexity. The complexity was mainly raised via larger filter size, more layers, and the value of strides in stage 1. After Stage 1, we added spatial dropout layers after convolutional layers with a stride of 2. Each feature extractor variant in Table~\ref{archetecture} is followed by an average pooling layer and a fully connected classification head with a Sigmoid output activation function.

\begin{table}[!htp]
\centering
\caption{Seed Designs of COVID-Net Assistant Models, Annotation Follows: "filter size,\ number of filters,\ stride" }
\label{archetecture}
\vspace{0.1cm}
\scalebox{0.9}{\begin{tabular}{c|c|c|c|c}\toprule
Seed Design &Stage 1 &Stage 2 &Stage 3 &Stage 4 \\\midrule
\multirow{3}{*}{CNN} &Conv 3x3, 32 &Conv 3x3, 64 &\multirow{3}{*}{Batch Normalization} &\multirow{3}{*}{-} \\
&Conv 3x3, 32 &Conv 3x3, 64 & & \\
&Maxpool 2x2 &Maxpool 2x2 & & \\
\hline 
\multirow{2}{*}{Res-CNN-S} &\multirow{2}{*}{Conv 3x3, 64, s2} &Res 3x3, 64, s2, d/o &\multirow{2}{*}{-} &\multirow{2}{*}{-} \\
& &Res 3x3, 64 & & \\
\hline 
\multirow{2}{*}{Res-CNN-M} &\multirow{2}{*}{Conv 3x3, 64, s2} &Res 3x3, 64, s2, d/o &Res 3x3, 64, s2, d/o &\multirow{2}{*}{-} \\
& &Res 3x3, 64 & Res 3x3, 64 & \\
\hline 
\multirow{2}{*}{Res-CNN-L} &\multirow{2}{*}{Conv 7x7, 64, s2} &Res 5x5, 64, s2, d/o &Res 3x3, 64, s2, d/o &Res 3x3, 64, s2, d/o \\
& &Res 5x5, 64 &Res 3x3, 64 &Res 3x3, 64 \\
\hline 
\multirow{2}{*}{DW-CNN-S} &\multirow{2}{*}{Conv 3x3, 32, s2} &DW 3x3, 64, s2, d/o &\multirow{2}{*}{DW 3x3, 128} &\multirow{2}{*}{-} \\
& &DW 3x3, 64 & & \\
\hline 
\multirow{2}{*}{DW-CNN-M} &\multirow{2}{*}{Conv 3x3, 32} &DW 3x3, 64, s2, d/o &DW 3x3, 128, s2, d/o &DW 3x3, 256, s2, d/o \\
& &DW 3x3, 64 &DW 3x3, 128 &DW 3x3, 256 \\
\hline 
\multirow{2}{*}{DW-CNN-L} &\multirow{2}{*}{Conv 9x9, 32} &DW 9x9, 64, s2, d/o &DW 7x7, 128, s2, d/o &DW 3x3, 256, s2, d/o \\
& &DW 9x9, 64 &DW 7x7, 128 &DW 3x3, 256 \\
\bottomrule
\end{tabular}}
\end{table}

\subsection{Training Policy}

All seed designs of proposed COVID-Net Assistant neural networks
were trained using the Adam optimizer with a binary cross
entropy loss function. In our training configuration, the learning rate would be decayed by a factor of 0.75 if the validation loss has not improved for two epochs, with an initial value of 0.0002 (ReduceLROnPlateau). The training would be terminated early if validation loss stopped improving for 10 epochs with 150 maximum training epochs (EarlyStopping). We constructed, trained, and
evaluated the models using TensorFlow Keras.

\section{Results}

We set up experiments to compare the representation power and efficiency of  COVID-Net Assistant models generated via generative synthesis~\cite{gensynth}, based on different seed designs introduced in section~\ref{model}. Specifically, we evaluated the Area under the Receiver Operating Characteristic Curve (AUC) as the primary performance metric. To express the theoretical capacity of models, we also calculated the total number of parameters and floating point operations (FLOPs) of the final architectures of COVID-Net Assistant models.

Table~\ref{result-theo} shows the performance and capacity of each proposed COVID-Net Assistant deep neural network generated via generative synthesis differentiated by seed designs. From Table~\ref{result-theo}, we noticed that COVID-Net Assistant CNN has dominant performance on the Verified-Only split, with the second smallest model in terms of the number of parameters. In addition, there is a slight increase in AUC by increasing the complexity of a residual block architected seed design (Res-CNN). In contrast, increasing the complexity of a depth-wise separable convolution architected seed design (DW-CNN) will cause a slight drop in the AUC score. The variants of COVID-Net Assistant DW-CNN have relatively small capacities, especially for models generated with seed design DW-CNN-S/M, whose numbers of FLOPs are less than 1/8 of other models. We also noticed a difference in the capacity levels of final architectures while the models were trained on different dataset splits with the same seed design. For example, the generated COVID-Net Res-CNN-S trained on All-Data Split has FLOPs fewer than half of the one trained on Verified-Only Split, while the number of parameters is only $3\%$ fewer. From our investigation, the difference is mainly in the number of filters in the first two stages described in Table~\ref{archetecture}, which are the most computationally expensive stages. We also noticed that a more complex seed design could result in a model with fewer FLOPs via generative synthesis. For instance, the final architecture of COVID-Net Assistant Res-CNN-M has $7\%$ fewer FLOPs than  COVID-Net Assistant Res-CNN-S, trained on the Verified-Only split, while Res-CNN-S has much fewer parameters.

\begin{table}[t]
\centering
\caption{Theoretical Evaluation Results of COVID-Net Assistant Models generated via Generative Synthesis based on Seed Designs, Format Follows "All-Data Split / Verified-Only Split", best result in \textbf{bold}}
\label{result-theo}
\vspace{0.1cm}
\begin{tabular}{c|c|c|c}\toprule

Seed Design &AUC &Params (K) &FLOPs (M) \\
\midrule
CNN &0.7815 / \textbf{0.9508} &6.6 / 4.6 &63.3 / 45.8 \\
Res-CNN-S &0.7828 / 0.9349  &74.3 / 76.3 &25.2 / 59.7 \\
Res-CNN-M &0.7887 / 0.9394  &161.4 / 228.8 &39.7 / 55.6 \\
Res-CNN-L  &\textbf{0.8039} / 0.9388 &254.8 / 229.0 &176 / 66.7 \\
DW-CNN-S &0.7860 / 0.9330  &\textbf{3.5} / \textbf{3.9} &\textbf{3.9\textbf} / \textbf{4.5} \\
DW-CNN-M &0.7832 / 0.9304 &25.2 / 23.7 &6.2 / 6.5 \\
DW-CNN-L &0.7692 / 0.9305 &34.1 / 35.2 &71.0 / 71.1 \\
\bottomrule
\end{tabular}
\end{table}

\begin{table}[t]
\centering
\caption{AUC Score of COVID-Net Assistant Models generated via Generative Synthesis based on Seed Designs Trained with All-Data Split on the Filtered Test Set, best result in \textbf{bold}}
\label{result-filtered}
\vspace{0.1cm}
\begin{tabular}{c|c}\toprule
Seed Design &AUC \\\midrule
CNN  &0.9411 \\
Res-CNN-S &0.9189 \\
Res-CNN-M &\textbf{0.9453} \\
Res-CNN-L  &0.9353 \\
DW-CNN-S &0.9439 \\
DW-CNN-M &0.9015 \\
DW-CNN-L  &0.9362 \\
\bottomrule
\end{tabular}
\end{table}

We noticed a considerable performance difference (over 0.1 AUC) on the test sets between the model trained on All-Data split and Verified-Only split. We hypothesized that the root cause is the more extensive noise of the unverified positive samples in All-Data split. To verify our hypothesis, we tested the models on the same test set in All-Data split but filtered \textrm{---} the unverified positive samples were rejected, left with 63 verified positive samples and 146 negative samples in the filtered test set. Table~\ref{result-filtered} shows the AUC scores on the filtered test set. As a result, the scores are close to the ones evaluated on Verfied-Only split, which is more than 10 points higher than the scores on the unfiltered test set, referred to Table~\ref{result-theo}. The observation implies that the unverified positive samples are noisy and may negatively impact the models trained via standard supervised learning.

Motivated by deploying the models on different platforms, such as mobile and web applications, we evaluated the model's forward pass latency on different platforms, reported in Table~\ref{result-latency}. Particularly, we computed the trimmed mean of the latency in 1000 forward propagation on 
\begin{enumerate*}[label={\arabic*)}]
	\item a workstation (WORKSTN) with x86\_64 architecture and Intel Xeon Gold 6230 CPU, 80 cores, 3.9 GHz, 400GB RAM;
	\item a single-board computer (SBC) with aarch64 architecture and Cortex-A72 CPU, 4 cores, 2.0 GHz, 4GB RAM.
\end{enumerate*} 
According to Table~\ref{result-latency}, a model's latency on the WORKSTN is not necessarily proportional to the latency on the SBC. For instance, the COVID-Net Assistant CNN trained on Verified-Only split can perform the fastest inference on the WORKSTN, but on the SBC, the latency of  COVID-Net Assistant DW-CNN-S is almost half of its. The models infer fastest on the SBC are COVID-Net Assistant DW-CNN-S/M, whose FLOPs are strictly fewer than other models. However, for lightweight models, The impact of fewer Flops on a workstation with abundant computational resources is not significant because of the large amount of parallel computation that can be performed in a deep convolutional neural network.  
\begin{table}[t]
\centering
\caption{Latency Evaluation Results of COVID-Net Assistant Models generated via Generative Synthesis based on Seed Designs, Format Follows "All-Data Split / Verified-Only Split", best result in \textbf{bold}}
\label{result-latency}
\vspace{0.1cm}
\begin{tabular}{c|c|c}\toprule

Seed Design &Latency WORKSTN (ms) &Latency SBC (ms) \\
\midrule
CNN  &2.4 / \textbf{1.72} &10.7 / 9.44 \\
Res-CNN-S &\textbf{2.29} / 2.22 &8.76 / 10.6 \\
Res-CNN-M  &2.64 / 2.61 &11.2 / 12.1  \\
Res-CNN-L  &3.24 / 3.05 &21.8 / 13.6  \\
DW-CNN-S  &2.34 / 2.31 &\textbf{4.94} / \textbf{5.39}  \\
DW-CNN-M  &3.06 / 3.18 &6.74 / 6.98  \\
DW-CNN-L  &14.0 / 14.1 &35.1 / 35.2  \\
\bottomrule
\end{tabular}
\end{table}
\section{Conclusion and Future Work}

In this study, we proposed the initiative of COVID-Net Assistant, a deep learning-driven system that performs pre-screening of COVID-19 conditions by processing cough audio. The goal of COVID-Net Assistant is not to make a diagnosis but to provide an early recommendation to users on whether they may have signs of COVID-19 and to seek medical advice or further COVID-19 tests. The study is purely research and should not be leveraged for medical advice or diagnosis. However, With effective pre-screening techniques and recommendation systems, limited medical sources like RT-PCR tests will gain more usability in the fight against the COVID-19 pandemic. 

Our results show that deep learning algorithms can do well on a dataset with verified positive samples. We proposed multiple deep convolutional neural network architectures generated via generative synthesis ~\cite{gensynth} with different seed architecture designs. We found the standard convolution based architecture, namely COVID-Net Assistant CNN, can achieve dominant representation power on the dataset with less noise; the lightweight depth-wise separable convolution based architecture, namely COVID-Net Assistant DW-CNN-S, can achieve the highest efficiency on low-end devices such as a single-board computer; and the residual block based architecture, namely COVID-Net Assistant Res-CNN-M, can achieve a great balance between representation power and efficiency, while the neural networks are not deep enough to fully leverage the power of residual connections.

One of the challenges is that the audio samples in the dataset are scarce. Due to the limited numbers of test data, the AUC scores with a slight difference are hardly comparable. More training data is necessary to generalize the model's target missions and deploy them in practice. In fact, deep learning algorithms have shown promising results in many diagnostic tasks with a massive amount of labeled training data. However,  high-quality labeled data in medical diagnosis is expensive to obtain and is always a challenge in related literature. Fortunately, recent studies have used semi-supervised learning approaches to leverage the unlabeled data or weak labels (e.g.,~\cite{ssl-covid}). As one of our future research directions, we will leverage transfer learning and semi-supervised learning algorithms to train the models on larger, crowdsourced datasets and leverage the unverified labels, such as combining the samples in COUGHVID~\cite{coughvid} dataset and Coswara~\cite{coswara} dataset with Covid19-Cough dataset~\cite{covid19-cough} currently in use.

The promising AUC scores on the Verified-Only dataset split also imply a potential pattern in coughing sounds by COVID-19 infection differentiated from other diseases. Hence, another research direction is leveraging explainable AI techniques to study the region of interest used in neural networks' prediction. This could further help us study and understand the performance of generative synthesis. We hope our research can motivate further study and applications or even research in biological reasoning. 

\section*{Acknowledgments}
We would like to thank DarwinAI Corporation, Vision and Image Processing Lab, and Saeejith Muralidharan Nair (University of Waterloo) for preparing the computing resources and platforms.  We would also like to thank the Natural Sciences and Engineering Research Council of Canada and the Canada Research Chairs program.

\bibliographystyle{IEEEtran}  
\bibliography{references.bib}

\begin{thebibliography}{10}
\providecommand{\url}[1]{#1}
\csname url@samestyle\endcsname
\providecommand{\newblock}{\relax}
\providecommand{\bibinfo}[2]{#2}
\providecommand{\BIBentrySTDinterwordspacing}{\spaceskip=0pt\relax}
\providecommand{\BIBentryALTinterwordstretchfactor}{4}
\providecommand{\BIBentryALTinterwordspacing}{\spaceskip=\fontdimen2\font plus
\BIBentryALTinterwordstretchfactor\fontdimen3\font minus
  \fontdimen4\font\relax}
\providecommand{\BIBforeignlanguage}[2]{{%
\expandafter\ifx\csname l@#1\endcsname\relax
\typeout{** WARNING: IEEEtran.bst: No hyphenation pattern has been}%
\typeout{** loaded for the language `#1'. Using the pattern for}%
\typeout{** the default language instead.}%
\else
\language=\csname l@#1\endcsname
\fi
#2}}
\providecommand{\BIBdecl}{\relax}
\BIBdecl

\bibitem{covidnet}
\BIBentryALTinterwordspacing
L.~Wang, Z.~Q. Lin, and A.~Wong, ``Covid-net: a tailored deep convolutional
  neural network design for detection of covid-19 cases from chest x-ray
  images,'' \emph{Scientific Reports}, vol.~10, no.~1, p. 19549, Nov 2020.
  [Online]. Available: \url{https://doi.org/10.1038/s41598-020-76550-z}
\BIBentrySTDinterwordspacing

\bibitem{covidnet-ct}
\BIBentryALTinterwordspacing
H.~Gunraj, L.~Wang, and A.~Wong, ``Covidnet-ct: A tailored deep convolutional
  neural network design for detection of covid-19 cases from chest ct images,''
  \emph{Frontiers in Medicine}, vol.~7, 2020. [Online]. Available:
  \url{https://www.frontiersin.org/articles/10.3389/fmed.2020.608525}
\BIBentrySTDinterwordspacing

\bibitem{coughvid}
\BIBentryALTinterwordspacing
L.~Orlandic, T.~Teijeiro, and D.~Atienza, ``The coughvid crowdsourcing dataset,
  a corpus for the study of large-scale cough analysis algorithms,''
  \emph{Scientific Data}, vol.~8, no.~1, p. 156, Jun 2021. [Online]. Available:
  \url{https://doi.org/10.1038/s41597-021-00937-4}
\BIBentrySTDinterwordspacing

\bibitem{coswara}
N.~Sharma, P.~Krishnan, R.~Kumar, S.~Ramoji, S.~Chetupalli, N.~R., P.~Ghosh,
  and S.~Ganapathy, ``Coswara — a database of breathing, cough, and voice
  sounds for covid-19 diagnosis,'' 10 2020, pp. 4811--4815.

\bibitem{covid19-cough}
\BIBentryALTinterwordspacing
P.~"fkthecovid", ``Dataset of recordings of induced cough.'' 2021. [Online].
  Available: \url{https://github.com/covid19-cough/dataset}
\BIBentrySTDinterwordspacing

\bibitem{librosa}
B.~McFee, C.~Raffel, D.~Liang, D.~P.~W. Ellis, M.~McVicar, E.~Battenberg, and
  O.~Nieto, ``librosa: Audio and music signal analysis in python,'' 2015.

\bibitem{relate-work2}
S.~K. Mahanta, D.~Kaushik, H.~Van~Truong, S.~Jain, and K.~Guha, ``Covid-19
  diagnosis from cough acoustics using convnets and data augmentation,'' in
  \emph{2021 First International Conference on Advances in Computing and Future
  Communication Technologies (ICACFCT)}, 2021, pp. 33--38.

\bibitem{relate-work1}
V.~Bansal, G.~Pahwa, and N.~Kannan, ``Cough classification for covid-19 based
  on audio mfcc features using convolutional neural networks,'' in \emph{2020
  IEEE International Conference on Computing, Power and Communication
  Technologies (GUCON)}, 2020, pp. 604--608.

\bibitem{gensynth}
A.~Wong, M.~J. Shafiee, B.~Chwyl, and F.~LI, ``Gensynth: A generative synthesis
  approach to learning generative machines to generate efficient neural
  networks,'' \emph{Electronics Letters}, vol.~55, 07 2019.

\bibitem{netscore}
A.~Wong, ``Netscore: Towards universal metrics for large-scale performance
  analysis of deep neural networks for practical usage,'' \emph{arXiv preprint
  arXiv:1806.05512}, vol.~4, 2018.

\bibitem{resnet}
K.~He, X.~Zhang, S.~Ren, and J.~Sun, ``Deep residual learning for image
  recognition,'' in \emph{2016 IEEE Conference on Computer Vision and Pattern
  Recognition (CVPR)}, 2016, pp. 770--778.

\bibitem{mobilenetv1}
A.~Howard, M.~Zhu, B.~Chen, D.~Kalenichenko, W.~Wang, T.~Weyand, M.~Andreetto,
  and H.~Adam, ``Mobilenets: Efficient convolutional neural networks for mobile
  vision applications,'' 04 2017.

\bibitem{res-beni}
A.~Veit, M.~Wilber, and S.~Belongie, ``Residual networks behave like ensembles
  of relatively shallow networks,'' in \emph{Proceedings of the 30th
  International Conference on Neural Information Processing Systems}, ser.
  NIPS'16.\hskip 1em plus 0.5em minus 0.4em\relax Red Hook, NY, USA: Curran
  Associates Inc., 2016, p. 550–558.

\bibitem{dw-conv}
\BIBentryALTinterwordspacing
L.~Kaiser, A.~N. Gomez, and F.~Chollet, ``Depthwise separable convolutions for
  neural machine translation,'' in \emph{International Conference on Learning
  Representations}, 2018. [Online]. Available:
  \url{https://openreview.net/forum?id=S1jBcueAb}
\BIBentrySTDinterwordspacing

\bibitem{ssl-covid}
Y.~Zhang, L.~Su, Z.~Liu, W.~Tan, Y.~Jiang, and C.~Cheng,
  ``\BIBforeignlanguage{en}{A semi-supervised learning approach for {COVID-19}
  detection from chest {CT} scans},''
  \emph{\BIBforeignlanguage{en}{Neurocomputing}}, vol. 503, pp. 314--324, Jun.
  2022.

\end{thebibliography}

\end{document}